\documentclass[10pt,twocolumn,letterpaper]{article}

\usepackage{wacv}              %

\usepackage{makecell}
\usepackage{multirow}

\definecolor{wacvblue}{rgb}{0.21,0.49,0.74}
\usepackage[pagebackref,breaklinks,colorlinks,allcolors=wacvblue]{hyperref}

\def\wacvPaperID{1420} %
\def\confName{WACV}
\def\confYear{2026}

\title{How Does Bilateral Ear Symmetry Affect Deep Ear Features?}

\author{\large Kagan Ozturk, Deeksha Arun, Kevin W. Bowyer, Patrick Flynn \\ \normalsize University of Notre Dame \\
}

\begin{document}
\maketitle

\def\thefootnote{}\footnotetext{Kagan Ozturk is  supported by the Ministry of National Education of Türkiye.}

\begin{abstract}

Ear recognition has gained attention as a reliable biometric technique due to the distinctive characteristics of human ears. With the increasing availability of large-scale datasets, convolutional neural networks (CNNs) have been widely adopted to learn features directly from raw ear images, outperforming traditional hand-crafted methods. However, the effect of bilateral ear symmetry on the features learned by CNNs has received little attention in recent studies. In this paper, we investigate how bilateral ear symmetry influences the effectiveness of CNN-based ear recognition. To this end, we first develop an ear side classifier to automatically categorize ear images as either left or right. We then explore the impact of incorporating this side information during both training and test. Cross-dataset evaluations are conducted on five datasets. Our results suggest that treating left and right ears separately during training and testing can lead to notable performance improvements. Furthermore, our ablation studies on alignment strategies, input sizes, and various hyperparameter settings provide practical insights into training CNN-based ear recognition systems on large-scale datasets to achieve higher verification rates.
\end{abstract}
    
\section{Introduction}
Biometric recognition systems have gained significant attention in recent years due to their applications in security, surveillance, and personal identification. Among various biometric traits, ear recognition has emerged as a promising alternative to traditional methods such as fingerprint, face, and iris recognition. The human ear possesses several advantageous characteristics that make it suitable for biometric identification: its structure remains relatively stable throughout a person's lifetime, it is not affected by facial expressions, and it can be captured from a distance. Moreover, the human body normally has two ears which exhibit similar (albeit mirrored) appearance, reflecting the bilateral symmetry inherent in the human body.

The uniqueness of ear morphology across individuals stems from its complex structure, which includes distinctive features such as the helix, antihelix, tragus, antitragus, concha, and the lobule. These anatomical elements form patterns that vary significantly between individuals, while maintaining consistency for each person over time. Furthermore, ear images can be acquired easily by using standard cameras, making ear recognition systems practical for real-world deployment.

Despite these advantages, ear recognition faces challenges such as occlusion by hair or accessories, varying illumination conditions, and different pose angles. Traditional approaches to ear recognition relied heavily on hand-crafted feature extraction methods, which often failed to capture the complex patterns necessary for high-accuracy identification. The advent of deep learning techniques, particularly CNNs, has revolutionized image recognition tasks by automatically learning features from raw data.

Face recognition research has progressed substantially, driven by the abundance of facial images available online \cite{cao2018vggface2, parkhi2015deep, schroff2015facenet, yi2014learning, zhu2021webface260m}. Although the primary intent of collecting these images is not to capture the ear region, a large proportion of these images inherently contain ear information that can be used to develop ear recognition systems. With the advent of reliable ear detection methods, new datasets dedicated to ear images have emerged, enabling more accurate recognition results compared to hand-crafted feature engineering approaches \cite{ramos2022vggfaceear, earvn, UERC2019, UERC2017, emersic2018convolutional, awe, benzaoui2023comprehensive, kamboj2021comprehensive}.

Although recent studies have explored the application of CNNs for ear recognition, this work focuses on three critical aspects that have received limited attention in the literature. First, we investigate the influence of ear symmetry on CNN-based feature extraction, an area that has not been studied rigorously in recent works. To this end, we develop a binary classifier to distinguish left and right ear images and analyze the impact of incorporating side information during both training and testing phases. Second, while the majority of existing works evaluate performance within the same dataset, without considering subject-disjoint train-test split, we conduct a comprehensive cross-dataset evaluation across five publicly available datasets, providing a more rigorous assessment of model generalization. Finally, we present an alignment strategy for improving recognition performance. Through extensive experiments with various hyperparameter settings, our analysis offers valuable insights into building more robust ear recognition systems.
\section{Related Work}
Ear recognition has emerged as an alternative to face recognition due to the robustness to face expressions and stability of the human ear. Consistent with recent trends in broader computer vision tasks, the methodologies for extracting discriminative features from ear images have shifted from hand-crafted descriptors to data-driven feature learning approaches. Comprehensive surveys of recent developments in ear biometrics can be found in~\cite{awe, benzaoui2023comprehensive, kamboj2021comprehensive, awex, galdamez2017brief, hassaballah2019ear}. 

One of the earliest attempts to benchmark ear recognition in unconstrained environments was the 2017 Unconstrained Ear Recognition Challenge (UERC) 2017~\cite{UERC2017}. The challenge provided a dataset comprising 11,804 images from 3,706 subjects. Various approaches were evaluated, including hand-crafted features such as Local Binary Patterns (LBP) and Chain Code Histogram (CCH), as well as deep learning architectures such as VGG~\cite{vggnet} and ResNet~\cite{resnet}. The best performance was achieved using CCH, with a Rank-1 recognition accuracy of $90.4\%$. Due to the limited size of the training set (2,304 images), deep learning methods underperformed, yielding less than $40\%$ accuracy.

The UERC was held again in 2019~\cite{UERC2019}, with a revised split comprising 1,500 training images and 10,000 test images. Compared to the 2017 challenge, only a modest improvement of $5\%$ in Rank-1 accuracy was observed, as the limited training data constrained the potential of deep learning methods.

A major breakthrough in large-scale ear image collection was introduced by Ramos et al.~\cite{ramos2022vggfaceear}, who leveraged an existing face dataset and an ear detection model to semi-automatically curate a new dataset (VGGFace-Ear) consisting of 234,651 ear images from 660 subjects. By finetuning a VGG-based network~\cite{vggnet}, they demonstrated that the performance of matchers based on feature learning methods significantly improves with the availability of larger training data.

The UERC 2023 challenge~\cite{UERC2023} has an expanded training with 248,655 images, including the VGGFace-Ear dataset~\cite{ramos2022vggfaceear} and additional newly collected samples. All participants in UERC 2023 employed deep learning-based feature extraction methods (one participant fused these with Histogram of Oriented Gradients features in a hybrid model). The highest overall recognition performance was achieved using a CNN-based feature learning approach.

\textbf{Bilateral ear symmetry} has been investigated in several studies in the context of ear recognition performance. One of the earliest works~\cite{yan2005empirical} reported that approximately $90\%$ of individuals have ears that are approximately bilaterally symmetric, although noticeable shape differences can exist between the left and right ears. In~\cite{abaza2010towards}, it was concluded that left and right ears are symmetric only to some extent. The authors reported a significant drop of $35\%$ in Rank-1 recognition accuracy when matching left ears against reflected right ears. The Unconstrained Ear Recognition Challenge (UERC2017) \cite{UERC2017} also evaluated accuracy for same-side and opposite-side ear matching to assess the impact of ear symmetry with hand-crafted features. Seven of the eight submitted techniques exhibited reduced performance with opposite-side matching while the best performing method show robustness to opposite-side ears.

\begin{figure*}
    \begin{center}
        \includegraphics[width=1\linewidth]{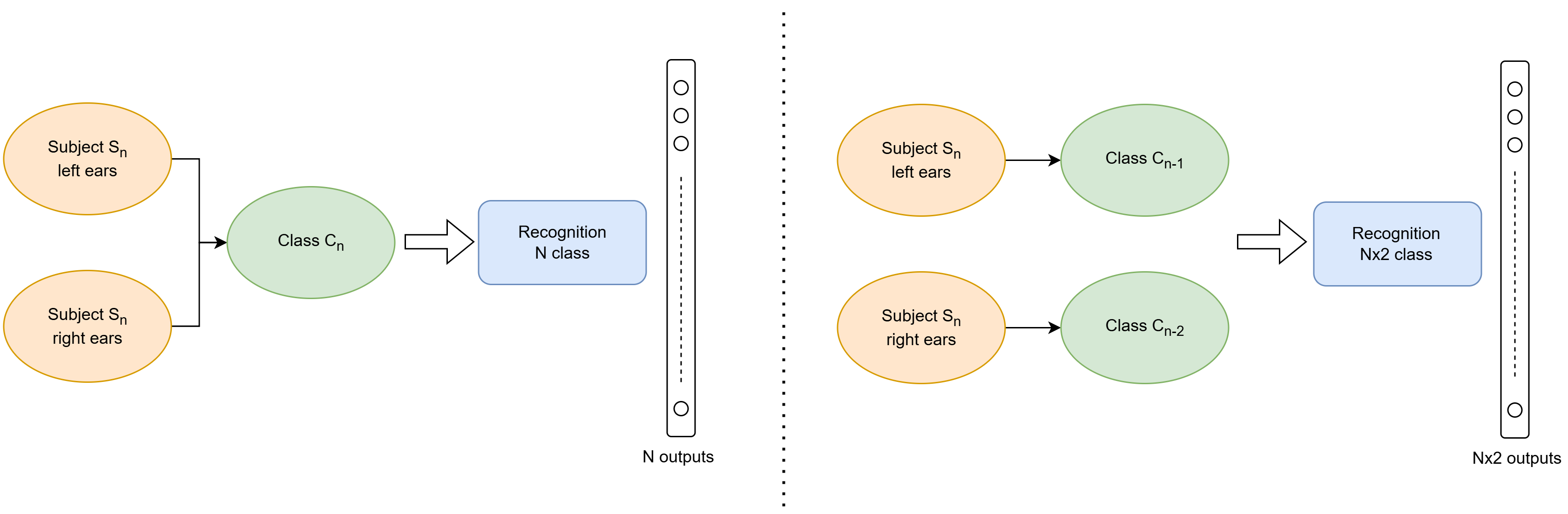}
    \end{center}
    \caption{Comparison of two training strategies to measure the impact of ear symmetry during training. First, training images are classified as left and right using an ear side classifier. Next, a CNN is trained for each approach: (i) Left and right ears of each subject are included in the same class, resulting in $N$ classes for $N$ subjects. (ii) Left and right ears are treated as different classes, resulting in $N \times2$ classes (assuming each subject in the training set has at least one left and right ear images).}
    \label{fig:training}
\end{figure*}

More recently, the impact of ear symmetry on deep learning-based features has been studied in~\cite{hansley2018employing, meng2021distinctiveness}. In~\cite{hansley2018employing}, they showed that matching same-side ears yields a lower Equal Error Rate (EER), with reductions of approximately $3\%$ on the AWE dataset and $5\%$ on the WPUT dataset. In~\cite{meng2021distinctiveness}, the authors reported $100\%$ recognition accuracy on their ear symmetry analysis, concluding that human ears are bilaterally symmetric. While aforementioned works analyzed the effect of ear symmetry, their findings regarding the role of symmetry in recognition performance remains questionable as limited number of images are used without considering cross-dataset evaluation settings.

The effect of \textbf{ear normalization} on recognition performance was studied in several works. Pflug \etal~\cite{pflug2014segmentation} introduced a geometrical normalization method using Cascaded Pose Regression (CPR) to rotate ear images until their principal axis was vertically aligned. Their results indicated that alignment prior to recognition consistently enhanced performance. Similarly, Hansley \etal~\cite{hansley2018employing} also observed improved recognition outcomes when alignment was applied before ear matching. Ribic \etal~\cite{ribivc2016influence} investigated the alignment effect using two subsets categorized by the severity of pose variations. Their first subset, which comprised images with only mild yaw and roll angles, demonstrated improved recognition accuracy with aligned ear images. However, in their second subset, which included images with large pose variations, alignment adversely affected recognition accuracy. Given that recognition rates were similar for unaligned images across both subsets, it can be claimed that advanced alignment techniques are necessary for effectively handling larger pose variations. More recently, Hrovatin \etal~\cite{hrovativc2023efficient} explored the impact of alignment on ear recognition using both hand-crafted and learning-based feature representations. They employed a two-stack hourglass model to detect 55 ear landmarks for alignment. Their findings confirmed the positive influence of alignment on recognition accuracy; however, the best-performing method achieved only a $30\%$ Rank-1 recognition rate on the AWEx dataset.

In this work, we explore several strategies to improve ear recognition performance in unconstrained environments. First, we propose a training method to examine how bilateral ear symmetry affects recognition accuracy when matching same-side and opposite-side ears. Next, we apply a segmentation-based ear alignment technique and assess its effectiveness on ear recognition. We evaluate the proposed approach using a cross-dataset evaluation protocol across four different datasets. Our study advances ear recognition research by demonstrating strong performance in cross-dataset evaluations, unlike recent studies that typically evaluate recognition accuracy using train-test splits within a single dataset.

\section{Methodology}
\subsection{Ear Side Classifier}
\label{sec:ear_side_classifier}
To analyze the impact of ear symmetry on ear recognition, we first develop an ear side classifier to categorize images as left and right. A CNN is employed for binary classification, using a modified ResNet architecture~\cite{resnet}. Specifically, the final layer of ResNet-100 is adapted for binary output, with a Global Average Pooling (GAP) layer applied after the last convolutional block to produce a 512-dimensional feature vector, followed by a linear layer to generate binary decision.

\textbf{Training.}
The network is trained using the CelebA-HQ dataset~\cite{celebahq}, where ear regions are cropped from face images based on hand-annotated landmarks provided with the CelebAMask-HQ dataset~\cite{celebahqmask}. To improve model robustness, data augmentation techniques including random horizontal and vertical translations ($[-5, 5]$ pixels), rotations ($[-15^\circ, 15^\circ]$), and random resized cropping ($\text{scale}=[0.5, 1]$, $\text{ratio}=[0.8, 1.3]$) are applied during training. All input images are resized to $64\times64$ pixels prior to being fed into the network.

A ResNet-100 backbone is used, as implemented in~\cite{insightface_github}, and dropout with a probability of 0.5 is applied after the GAP layer to enhance generalization. The model is optimized using stochastic gradient descent (SGD) with binary cross-entropy loss over 30 epochs. For cross-dataset evaluation, performance is assessed on the AWE~\cite{awe} dataset. During inference, images are resized to $96\times96$ pixels to improve prediction accuracy~\cite{touvron2019fixing}.

\subsection{Alignment}
\label{sec:alignment}

\begin{figure}[b]
    \begin{center}
        \includegraphics[width=1\linewidth]{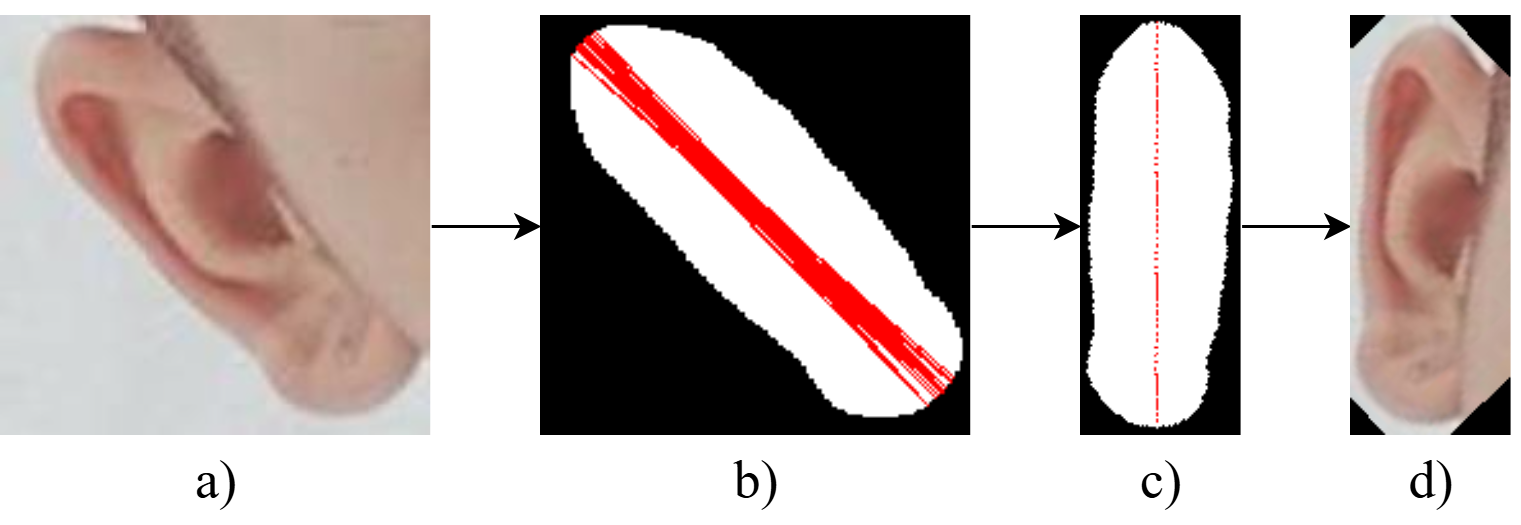}
    \end{center}
    \caption{Proposed alignment strategy. a) Process images to detect ear region using a segmentation model. b) Then find the longest top-$k$ lines in the segmented region and calculate the mean top and bottom point of the ear. c) Rotate the image for alignment of the line from top to bottom. d) Crop the segmented region.}
    \label{fig:alignment}
\end{figure}

\begin{figure}
    \begin{center}
        \includegraphics[width=1\linewidth]{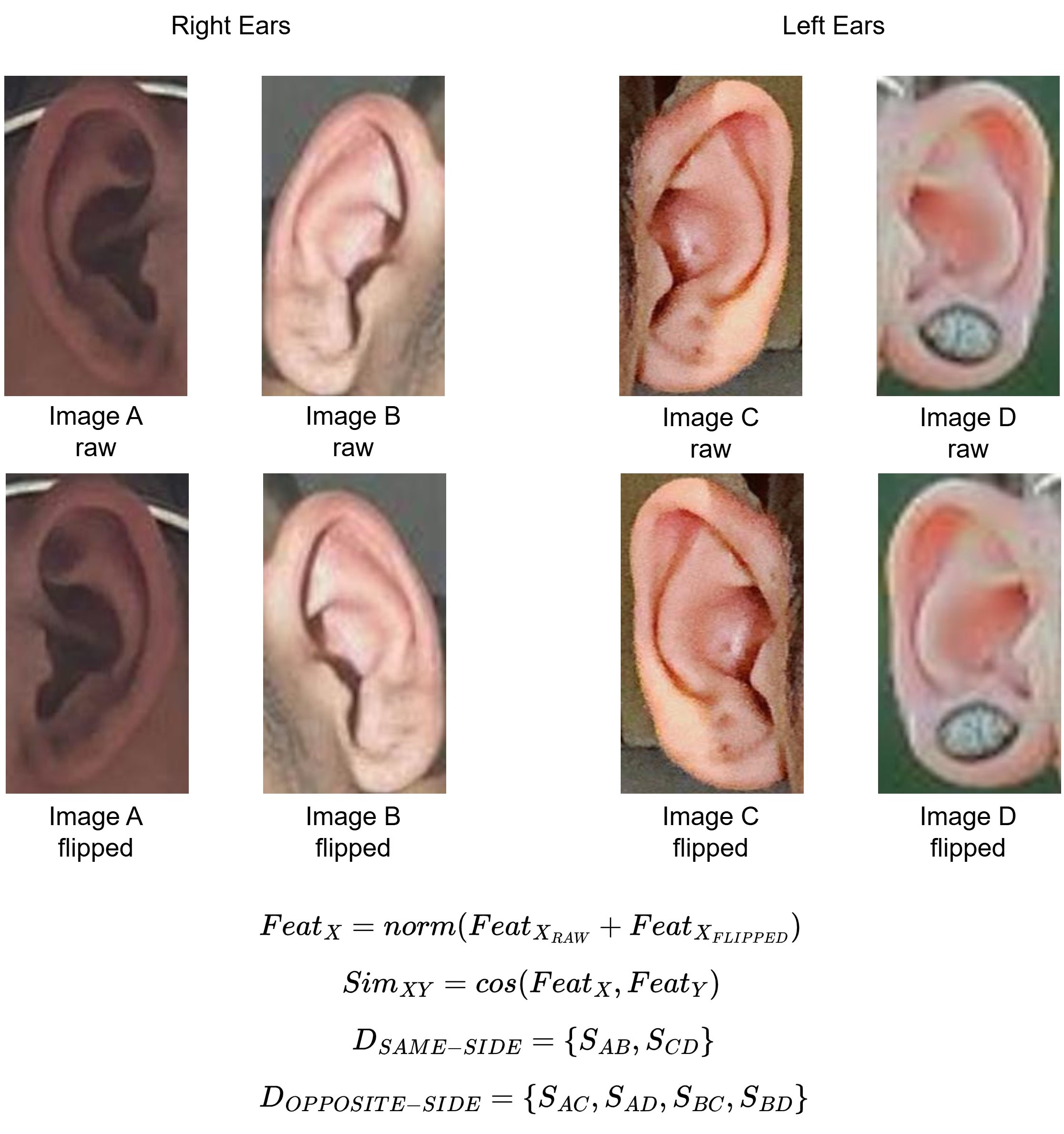}
    \end{center}
    \caption{Test images are first classified as left or right using an ear-side classifier. Images are horizontally flipped for test time augmentation. We compute similarity score distributions for two settings: matching same-side ears ($D_{\text{SAME-SIDE}}$) and opposite-side ears ($D_{\text{OPPOSITE-SIDE}}$), to assess left-right ear similarity. Verification results are reported in Table~\ref{tab:side}.}
    \label{fig:test}
\end{figure}

We propose an alignment strategy to assess its effect for ear recognition. An overview of the alignment method can be seen in Figure~\ref{fig:alignment}. First, we train a segmentation model to detect ear pixels. Next, bottom and top points of ears are calculated by finding the mean of the top-$k$ longest lines.  The number of $k$ lines are set empirically to increase robustness of the predictions. Then, images are rotated according to the line and the segmented region is cropped to obtain aligned images.

\textbf{Training.} 
We use cropped ear images from the CelebA-HQ dataset~\cite{celebahq, celebahqmask} to train the ear segmentation model. The BiSeNet architecture~\cite{bisenet} is utilized, based on the publicly available implementation provided in~\cite{bisenet_github}. To enhance model generalization, data augmentation techniques including random color jittering (with brightness, contrast, and saturation factors sampled from $[0, 0.5]$) and JPEG compression (with quality factors sampled from $[10, 90]$) are applied during training. Optimization is performed using the Adam optimizer. The model is trained for 50 epochs. Input images are resized to $224\times128$   for input to the network.

\subsection{Ear Recognition}
\label{sec:ear_recognition}
This work investigates the influence of left-right (approximate) ear similarity on CNN-based ear representations in the training and inference phases. Two training strategies are explored: (i) single class per subject, where left and right ear images are grouped together, resulting in $N$ classes for $N$ subjects, and (ii) separate classes for each ear side, resulting in $N\times2$ classes for $N$ subjects. For both strategies, the ResNet-100 backbone is utilized to train the ear recognition models.

\textbf{Training.}
Following the setup in~\cite{arcface}, a ResNet-100 architecture is employed with the ArcFace loss to learn ear representations. The training is conducted using images from the UERC 2019 dataset~\cite{UERC2019}, with cross-dataset evaluations performed on four benchmark datasets: AWE~\cite{awe}, AWEx~\cite{awex}, WPUT~\cite{wput} and EarVN~\cite{earvn}. An overview of the two training strategies designed to assess the effect of ear symmetry is illustrated in Figure~\ref{fig:training}. Additionally, we analyze the impact of the proposed alignment method and different input resolutions on recognition performance.

We adopt the publicly available ArcFace implementation in~\cite{insightface_github} to train our models. Hyperparameter optimization is performed for learning rate and dropout rate. When training the network with randomly initialized weights, a learning rate of $0.5$ is used. For finetuning, we initialize the network with weights from a ResNet-100 model pretrained for face recognition~\cite{arcface} on the WebFace4M dataset~\cite{zhu2021webface260m}. In this finetuning setting, a learning rate of $0.01$ is used to train all layers of the network. A dropout rate of $0.8$ is applied after the fully connected layers for regularization, as higher dropout was found to be effective in our preliminary experiments.

\textbf{Inference.} Once the training is done, two strategies are employed to evaluate the effect of ear symmetry on the test set. An overview of the protocol can be seen in Figure~\ref{fig:test}. First, ear images are categorized as left and right using the ear side classifier. For all images, horizontal flipping is applied to augment test images. By doing so, we not only increase verification results but also ensure that our analysis on left and right ear similarity is not influenced by the opposite orientation. For an image $X$, feature representation is obtained by normalizing the sum of the features of raw and flipped images $Feat_X = norm(Feat_{X_{RAW}}+Feat_{X_{FLIPPED}})$. Then, the cosine similarity score is computed: $Sim_{XY} = cos(Feat_X, Feat_Y)$.

\section{Experiments}

\subsection{Datasets}

\begin{figure}
    \begin{center}
        \includegraphics[width=0.9\linewidth]{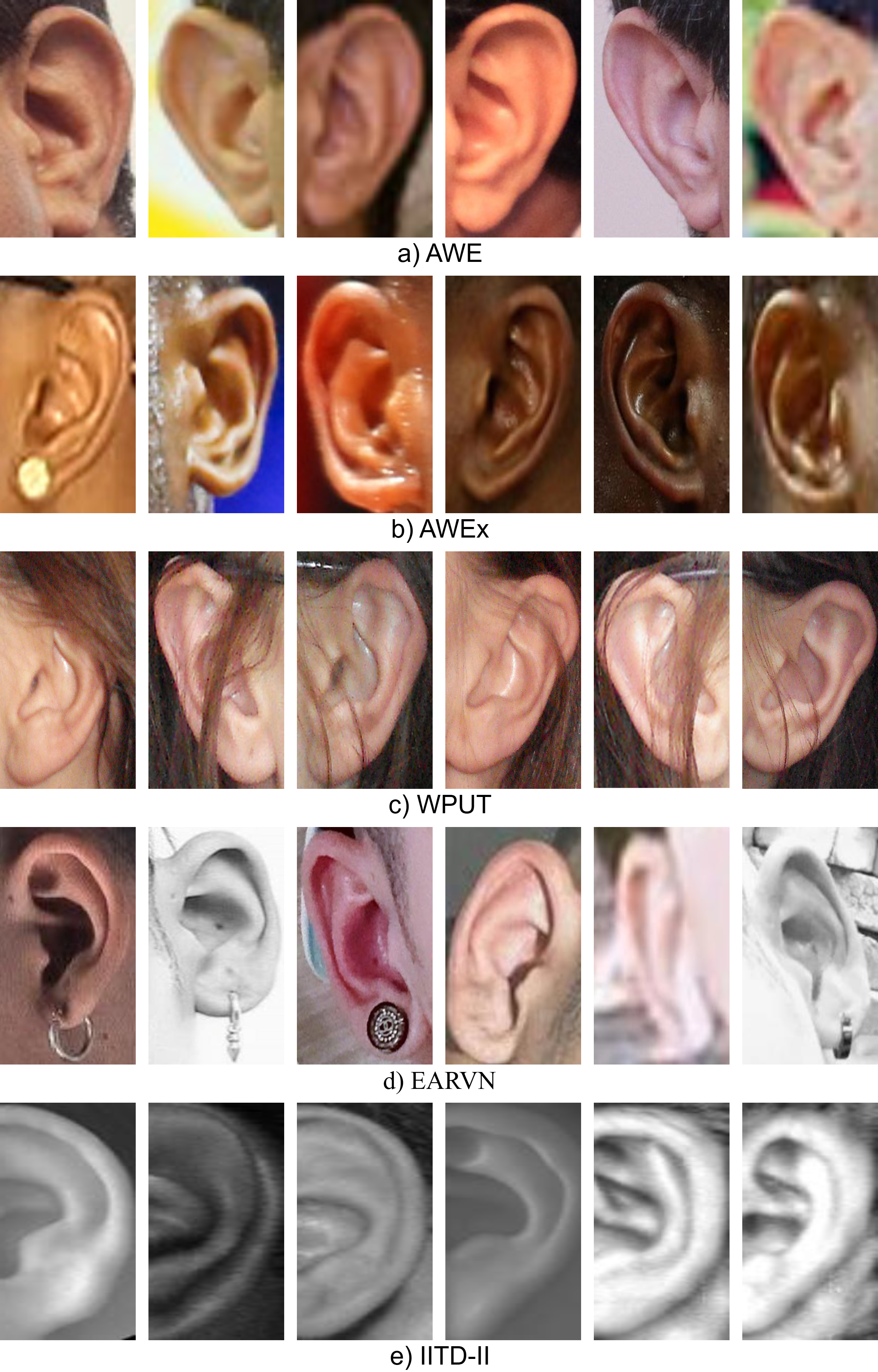}
    \end{center}
    \caption{Example images from each of the five datasets used for evaluating recognition performance.}
    \label{fig:datasets}
\end{figure}

We use seven datasets in our experiments. The CelebA-HQ dataset is used to train the ear side classifier and the ear segmentation model. The UERC2023 dataset is used to train the ear recognition models. Recognition performance is evaluated on four benchmark datasets: AWE, AWEx, WPUT, and EarVN. Example images from the evaluation sets are shown in Figure~\ref{fig:datasets}.

\textbf{CelebA-HQ} dataset~\cite{celebahq} is derived from the CelebA dataset \cite{celeba} to create high resolution versions of 30,000 images. They apply several image processing steps to ensure consistent quality as CelebA images vary significantly in resolution. First, JPEG artifacts are removed using an auto-encoder. Next, a super-resolution network is used to enhance quality. Images are then cropped and aligned using facial landmarks. Finally, all images are resized to $1024\times1024$.

Manual annotations for CelebA-HQ images are provided by the CelebAMask-HQ dataset~\cite{celebahqmask}. The annotation contains 19 classes including facial parts such as, `nose', `eye' and `ear'. Out of the 30,000 images, 14,323 right ears and 15,865 left ears are annotated. In our experiments, we use tightly cropped ear regions for training the ear side classifier and the ear segmentation model.

\textbf{UERC 2023}~\cite{UERC2023} dataset is used to train our ear recognition models. The training set comprises of 248,655 images from 1,310 subjects. For our experiments, we exclude the first 100 subjects, corresponding to the AWE dataset. Thus, our model is trained on 247,655 images from 1,210 subjects.

The \textbf{AWE} dataset~\cite{awe} is one of the earliest attempts to collect ear images captured in unconstrained environments. It includes 1,000 images from 100 subjects, with each subject contributing 10 images of varying quality and resolution.
It also includes human-annotated metadata for the following categories: gender, ethnicity, accessories, occlusion, head pitch, head roll, head yaw, and ear side.

\textbf{AWEx} dataset~\cite{awex} is an extended version of the AWE dataset with 220 new subjects. The same protocol is followed to acquire ear images from the web. Each subject has 10 images.

\textbf{WPUT} dataset~\cite{wput} includes 2,071 images from 501 subjects collected in a constrained setting. For each subject, at least 4 images are provided, capturing both left and right ears at profile (90°) and half-profile (75°) angles. The dataset introduces significant variation in occlusions such as hair, earrings, and glasses. Each image is accompanied by  metadata describing subject demographics, capture conditions, and occlusion types.

\textbf{EarVN1.0} dataset~\cite{earvn} is the largest ear datasets in our evaluation set. It consists of 28,412 images of 164 Asian subjects. Each subject has at least 100 images. The dataset exhibits great variation in pose, illumination and resolution. While it is commonly included in both training and test sets for evaluating the performance of the recent ear recognition models, we use all images in the test set to provide cross-dataset evaluation. As the dataset includes greater variation in pose and resolution, compared to the other three benchmark datasets, it is the most challenging dataset for performance evaluation.

\textbf{IITD-II} dataset~\cite{kumar2012automated} is comprised of 793 images from 221 subjects. It is used for performance comparison with previous work.

\subsection{Results}

In this study, we propose two complementary models to enhance the performance of ear recognition systems. First, we develop an ear side classifier trained on the CelebA-HQ dataset and evaluate its performance on the AWE dataset, which contains 520 left and 480 right ear images. The binary classifier achieves an accuracy of $99.8\%$ on this task. Second, we introduce an ear alignment strategy based on an ear segmentation model, also trained on the CelebA-HQ dataset. The effectiveness of the proposed alignment method is assessed through improvements in ear recognition performance metrics.

\setlength{\tabcolsep}{5pt} 
\renewcommand{\arraystretch}{1.0}
\begin{table*}[hbt]
\begin{center}
\begin{tabular}{|c|c||c|c|c|c||c|c|c|c|}
\hline

\multicolumn{2}{|c||}{} & \multicolumn{4}{c||}{$AUC\uparrow$} & \multicolumn{4}{c|}{$FNMR@FMR=1\%\downarrow$} \\
\hline

\makecell{Input size \\ (height $\times$ width)} & Finetuning & AWE & AWEx & WPUT & EARVN & AWE & AWEx & WPUT & EARVN \\
\hline

\multirow{2}{*}{$112 \times 112$} & \multirow{2}{*}{No} 
& $0.997$ & $0.988$ & $0.980$ & $0.927$ 
& $0.027$ & $0.100$ & $0.145$ & $0.425$ \\
& & $\pm 6.8e\text{-}4$ & $\pm 9.0e\text{-}4$ & $\pm 1.8e\text{-}3$ & $\pm 7.3e\text{-}4$ & $\pm 3.2e\text{-}3$ & $\pm 6.7e\text{-}3$ & $\pm 9.5e\text{-}3$ & $\pm 2.8e\text{-}3$ \\
\hline

\multirow{2}{*}{$192 \times 112$} & \multirow{2}{*}{No} 
& $0.997$ & $0.990$ & $0.990$ & $0.950$ 
& $0.022$ & $0.089$ & $0.089$ & $0.315$ \\
& & $\pm 3.6e\text{-}4$ & $\pm 4.6e\text{-}4$ & $\pm 1.1e\text{-}3$ & $\pm 8.2e\text{-}4$ & $\pm 2.6e\text{-}3$ & $\pm 2.5e\text{-}3$ & $\pm 4.8e\text{-}3$ & $\pm 2.4e\text{-}3$ \\
\hline

\multirow{2}{*}{$224 \times 128$} & \multirow{2}{*}{No} 
& $0.998$ & $0.989$ & $0.990$ & $0.953$ 
& $0.018$ & $0.085$ & $0.078$ & $0.296$ \\
& & $\pm 3.9e\text{-}4$ & $\pm 7.4e\text{-}4$ & $\pm 6.3e\text{-}4$ & $\pm 6.4e\text{-}4$ & $\pm 2.0e\text{-}3$ & $\pm 2.9e\text{-}3$ & $\pm 3.1e\text{-}3$ & $\pm 2.3e\text{-}3$ \\
\hline

\multirow{2}{*}{$224 \times 224$} & \multirow{2}{*}{No} 
& $0.997$ & $0.990$ & $0.991$ & $0.951$ 
& $0.020$ & $0.085$ & $0.084$ & $0.288$ \\
& & $\pm 7.3e\text{-}4$ & $\pm 7.2e\text{-}4$ & $\pm 8.4e\text{-}4$ & $\pm 5.9e\text{-}4$ & $\pm 2.2e\text{-}3$ & $\pm 6.3e\text{-}3$ & $\pm 6.2e\text{-}3$ & $\pm 2.4e\text{-}3$ \\
\hline

\multirow{2}{*}{$112 \times 112$} & \multirow{2}{*}{Yes} 
& \textcolor{blue}{$0.999$} & $0.992$ & $0.990$ & $0.940$ 
& $0.017$ & \textcolor{blue}{$0.071$} & $0.088$ & $0.373$ \\
& & $\pm 1.8e\text{-}4$ & $\pm 7.8e\text{-}4$ & $\pm 1.2e\text{-}3$ & $\pm 5.4e\text{-}4$ & $\pm 2.2e\text{-}3$ & $\pm 2.5e\text{-}3$ & $\pm 5.1e\text{-}3$ & $\pm 2.5e\text{-}3$ \\
\hline

\multirow{2}{*}{$192 \times 112$} & \multirow{2}{*}{Yes} 
& $0.999$ & \textcolor{blue}{$0.992$} & $0.992$ & $0.953$ 
& \textcolor{blue}{$0.012$} & $0.072$ & $0.078$ & $0.301$ \\
& & $\pm 1.6e\text{-}4$ & $\pm 5.2e\text{-}4$ & $\pm 1.5e\text{-}3$ & $\pm 4.3e\text{-}4$ & $\pm 2.3e\text{-}3$ & $\pm 5.0e\text{-}3$ & $\pm 5.7e\text{-}3$ & $\pm 2.1e\text{-}3$ \\
\hline

\multirow{2}{*}{$224 \times 128$} & \multirow{2}{*}{Yes} 
& $0.998$ & $0.992$ & \textcolor{blue}{$0.993$} & \textcolor{blue}{$0.956$} 
& $0.018$ & $0.075$ & \textcolor{blue}{$0.073$} & \textcolor{blue}{$0.280$} \\
& & $\pm 2.7e\text{-}4$ & $\pm 6.4e\text{-}4$ & $\pm 9.7e\text{-}4$ & $\pm 5.9e\text{-}4$ & $\pm 2.9e\text{-}3$ & $\pm 4.6e\text{-}3$ & $\pm 8.3e\text{-}3$ & $\pm 1.9e\text{-}3$ \\
\hline

\multirow{2}{*}{$224 \times 224$} & \multirow{2}{*}{Yes} 
& $0.998$ & $0.990$ & $0.991$ & $0.952$ 
& $0.020$ & $0.089$ & $0.086$ & $0.292$ \\
& & $\pm 2.9e\text{-}4$ & $\pm 4.4e\text{-}4$ & $\pm 7.2e\text{-}4$ & $\pm 7.0e\text{-}4$ & $\pm 2.7e\text{-}3$ & $\pm 3.9e\text{-}3$ & $\pm 5.4e\text{-}3$ & $\pm 2.6e\text{-}3$ \\
\hline

\end{tabular}
\end{center}
\caption{Ablation study on the effect of input resolution and finetuning a face recognition model on ear recognition performance. For each dataset, the highest AUC and lowest FNMR values are highlighted in blue.}
\label{tab:size_and_finetune}
\end{table*}

\setlength{\tabcolsep}{5pt} 
\renewcommand{\arraystretch}{1.0}
\begin{table*}[hbt]
\begin{center}
\begin{tabular}{|c|c||c|c|c|c||c|c|c|c|}
\hline

\multicolumn{2}{|c||}{Alignment} & \multicolumn{4}{c||}{$AUC\uparrow$} & \multicolumn{4}{c|}{$FNMR@FMR=1\%\downarrow$} \\
\hline

Training & Test & AWE & AWEx & WPUT & EARVN & AWE & AWEx & WPUT & EARVN \\
\hline

\multirow{2}{*}{No} & \multirow{2}{*}{No} 
& \textcolor{blue}{$0.999$} & $0.992$ & \textcolor{blue}{$0.996$} & $0.928$ 
& \textcolor{blue}{$0.011$} & \textcolor{blue}{$0.072$} & \textcolor{blue}{$0.058$} & $0.435$ \\
& & $\pm 3.1e\text{-}4$ & $\pm 5.7e\text{-}4$ & $\pm 4.7e\text{-}4$ & $\pm 3.6e\text{-}3$ & $\pm 1.7e\text{-}3$ & $\pm 3.6e\text{-}3$ & $\pm 4.9e\text{-}3$ & $\pm 1.0e\text{-}3$ \\
\hline

\multirow{2}{*}{No} & \multirow{2}{*}{Yes} 
& $0.998$ & $0.986$ & $0.992$ & $0.939$ 
& $0.027$ & $0.126$ & $0.080$ & $0.365$ \\
& & $\pm 4.2e\text{-}4$ & $\pm 8.1e\text{-}4$ & $\pm 1.0e\text{-}3$ & $\pm 2.0e\text{-}4$ & $\pm 2.5e\text{-}3$ & $\pm 5.0e\text{-}3$ & $\pm 5.6e\text{-}3$ & $\pm 2.0e\text{-}3$ \\
\hline

\multirow{2}{*}{Yes} & \multirow{2}{*}{No} 
& $0.995$ & $0.976$ & $0.983$ & $0.886$ 
& $0.062$ & $0.186$ & $0.176$ & $0.549$ \\
& & $\pm 6.8e\text{-}4$ & $\pm 1.0e\text{-}3$ & $\pm 1.4e\text{-}3$ & $\pm 1.1e\text{-}3$ & $\pm 6.5e\text{-}3$ & $\pm 6.7e\text{-}3$ & $\pm 1.1e\text{-}2$ & $\pm 2.5e\text{-}3$ \\
\hline

\multirow{2}{*}{Yes} & \multirow{2}{*}{Yes} 
& $0.998$ & \textcolor{blue}{$0.992$} & $0.992$ & \textcolor{blue}{$0.956$} 
& $0.018$ & $0.075$ & $0.073$ & \textcolor{blue}{$0.280$} \\
& & $\pm 2.7e\text{-}4$ & $\pm 6.4e\text{-}4$ & $\pm 9.7e\text{-}4$ & $\pm 5.9e\text{-}4$ & $\pm 2.9e\text{-}3$ & $\pm 4.6e\text{-}3$ & $\pm 8.3e\text{-}3$ & $\pm 1.9e\text{-}3$ \\
\hline

\end{tabular}
\end{center}
\caption{Ablation study on the effect of alignment. For each dataset, the highest AUC and lowest FNMR values are highlighted in blue.}
\label{tab:align}
\end{table*}

\setlength{\tabcolsep}{5pt} 
\renewcommand{\arraystretch}{1.0}
\begin{table*}[hbt]
\begin{center}
\begin{tabular}{|c|c||c|c|c|c||c|c|c|c|}
\hline

\multicolumn{2}{|c||}{Left-Right Separation} & \multicolumn{4}{c||}{$AUC\uparrow$ } & \multicolumn{4}{c|}{$FNMR@FMR=1\%\downarrow$} \\
\hline

Training & Test & AWE & AWEx & WPUT & EARVN & AWE & AWEx & WPUT & EARVN \\
\hline

\multirow{2}{*}{No} & \multirow{2}{*}{No (Opposite side)} 
& $0.994$ & $0.985$ & $0.981$ & $0.931$ 
& $0.094$ & $0.154$ & $0.199$ & $0.447$ \\
& & $\pm 1.2e\text{-}3$ & $\pm 1.4e\text{-}3$ & $\pm 1.0e\text{-}3$ & $\pm 7.2e\text{-}4$ & $\pm 1.1e\text{-}2$ & $\pm 9.3e\text{-}3$ & $\pm 7.0e\text{-}3$ & $\pm 2.0e\text{-}3$ \\
\hline

\multirow{2}{*}{No} & \multirow{2}{*}{Yes (Same Side)} 
& \textcolor{blue}{$0.999$} & \textcolor{blue}{$0.993$} & $0.991$ & $0.949$ 
& $0.023$ & \textcolor{blue}{$0.066$} & $0.075$ & $0.336$ \\
& & $\pm 2.7e\text{-}4$ & $\pm 4.6e\text{-}4$ & $\pm 1.1e\text{-}3$ & $\pm 6.6e\text{-}4$ & $\pm 3.4e\text{-}3$ & $\pm 2.1e\text{-}3$ & $\pm 5.5e\text{-}3$ & $\pm 2.3e\text{-}3$ \\
\hline

\multirow{2}{*}{Yes} & \multirow{2}{*}{No (Opposite Side)} 
& $0.988$ & $0.976$ & $0.977$ & $0.931$ 
& $0.159$ & $0.241$ & $0.246$ & $0.432$ \\
& & $\pm 2.1e\text{-}3$ & $\pm 1.9e\text{-}3$ & $\pm 1.3e\text{-}3$ & $\pm 7.5e\text{-}4$ & $\pm 1.1e\text{-}2$ & $\pm 5.5e\text{-}3$ & $\pm 7.8e\text{-}3$ & $\pm 1.7e\text{-}3$ \\
\hline

\multirow{2}{*}{Yes} & \multirow{2}{*}{Yes (Same Side)} 
& $0.998$ & $0.992$ & \textcolor{blue}{$0.993$} & \textcolor{blue}{$0.956$} 
& \textcolor{blue}{$0.018$} & $0.075$ & \textcolor{blue}{$0.073$} & \textcolor{blue}{$0.280$} \\
& & $\pm 2.7e\text{-}4$ & $\pm 6.4e\text{-}4$ & $\pm 9.7e\text{-}4$ & $\pm 5.9e\text{-}4$ & $\pm 2.9e\text{-}3$ & $\pm 4.6e\text{-}3$ & $\pm 8.3e\text{-}3$ & $\pm 1.9e\text{-}3$ \\
\hline

\end{tabular}
\end{center}
\caption{Ablation study on evaluating the impact of left-right ear separation during training and testing. In the ``Training" column, ``No" indicates models trained with $N$ classes (combining left and right ear images as one class per subject), while ``Yes" refers to models trained with $N \times 2$ classes (separating left and right ears as two different classes), as described in Section \ref{sec:ear_recognition}. Two testing protocols are compared: matching opposite-side ears (left-to-right) and matching same-side ears (left-to-left and right-to-right). For each dataset, the highest AUC and lowest FNMR values are highlighted in blue.}
\label{tab:side}
\end{table*}

\setlength{\tabcolsep}{5pt} 
\renewcommand{\arraystretch}{1.2}
\begin{table}[b]
\begin{center}
\begin{tabular}{|c|c|c|}
\hline

\multirow{2}{*}{Method} & \multicolumn{1}{c|}{IITD-II} & \multirow{2}{*}{\makecell{Cross-dataset \\ Evaluation}} \\

 & \multicolumn{1}{c|}{Rank-1$\uparrow$} & \\
\hline

Kumar et al. \cite{kumar2012automated} & $95.93$ & No\\
Dodge et al. \cite{dodge2018unconstrained} & $92.40$ & No\\
Meraoumia et al. \cite{meraoumia2015automated} & $95.20$ & No \\
Zarachoff et al. \cite{zarachoff2021non} & $94.47$ & No \\
Chang et al. \cite{chang2003comparison} & $89.78$ & No \\
Ahila et al. \cite{ahila2021deep} & $97.36$ & No \\
Mehta et al. \cite{mehta2024efficient} & $\mathbf{98.74}$ & No \\
 \hline
 
\textbf{Ours} & $\mathbf{97.73}$ & \textbf{Yes} \\

\hline

\end{tabular}
\end{center}
\caption{Performance comparison with previous work.}
\label{tab:sota}
\end{table}

\textbf{Ear Recognition Performance.} We use the cosine similarity between two 512-dimensional ear representations to calculate scores for 1:1 comparisons. Four datasets are used for cross-dataset evaluation. Two metrics are used for reporting the verification performance: Area Under the Curve (AUC) and False Non-Match Rate (FNMR) at a $1\%$ False Match Rate (FMR).

We begin our experiments by investigating the effect of input image resolution on recognition performance. Additionally, we evaluate the impact of finetuning a pretrained face recognition model on the ear recognition task. Various input resolutions with different aspect ratios are considered: $112\times112$, $192\times112$, $224\times128$, and $224\times224$. The results are reported in Table~\ref{tab:size_and_finetune}. Note that while all models are trained using the same ResNet-100 architecture, the number of parameters increases with higher input resolutions due to the fully connected layer at the end of the network.

We observe that initializing the ear recognition network with the weights of the pretrained face recognition network improves the results in almost all cases. As shown in Table~\ref{tab:size_and_finetune}, this improvement is particularly notable when the input resolution is close to $112\times112$, which matches the resolution used during pretraining of the face recognition model. The performance gap between models initialized randomly and those finetuned from pretrained weights narrows as the input resolution increases. Nevertheless, larger input sizes generally lead to better performance, with or without finetuning, especially on the more challenging WPUT and EarVN datasets. Furthermore, the results indicate that using input resolutions with greater height than width tends to yield better performance compared to square dimensions. As we observe the best performance with $224\times128$ on average, we use this resolution with finetuning approach for the rest of our experiments.

\begin{figure*}
    \begin{subfigure}[b]{0.49\textwidth}
        \includegraphics[width=\textwidth]{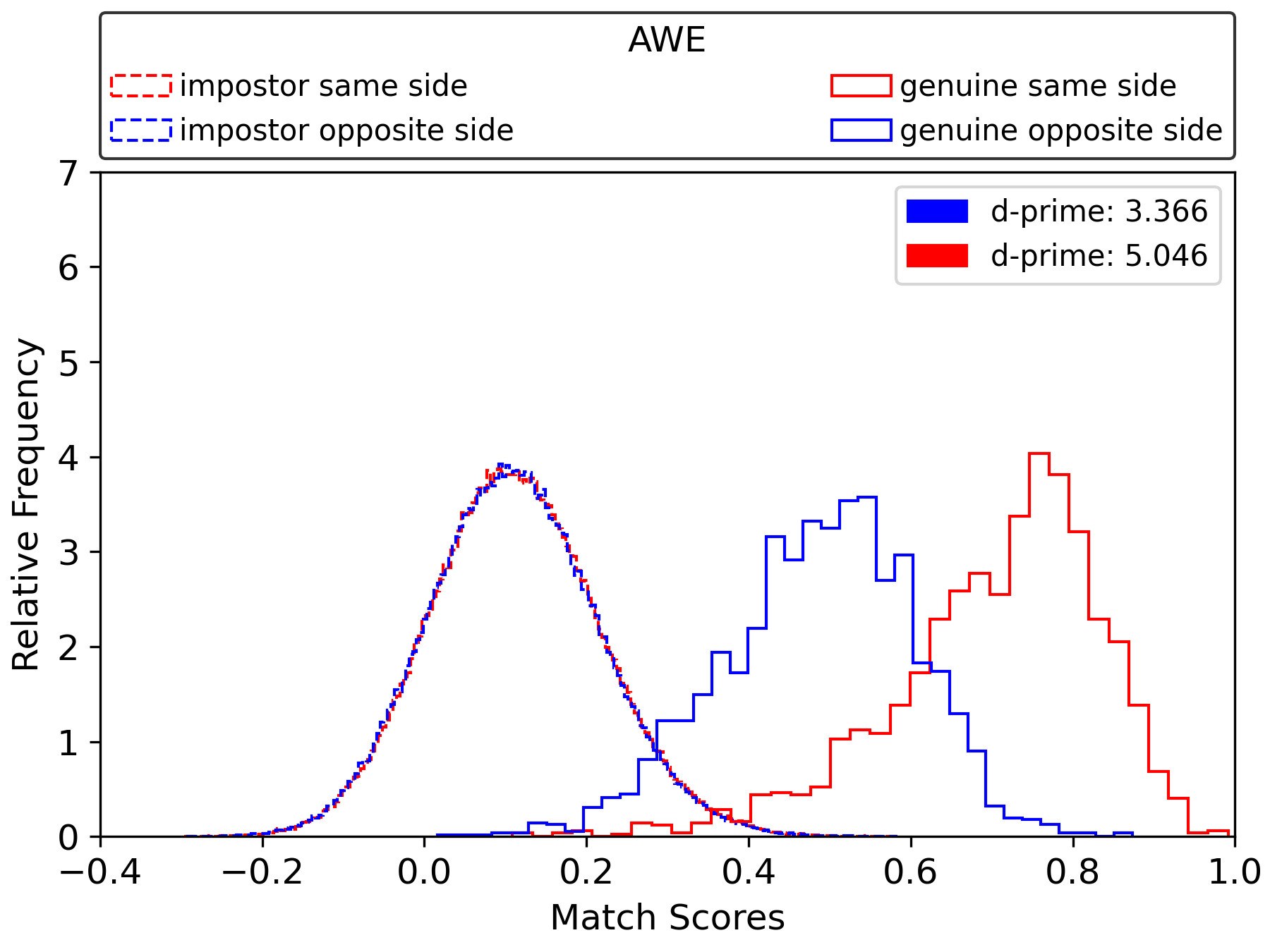}
        \label{fig:awe}
        \vspace{-10pt}
        \caption{AWE}
    \end{subfigure}
    \hfill
    \begin{subfigure}[b]{0.49\textwidth}
        \includegraphics[width=\textwidth]{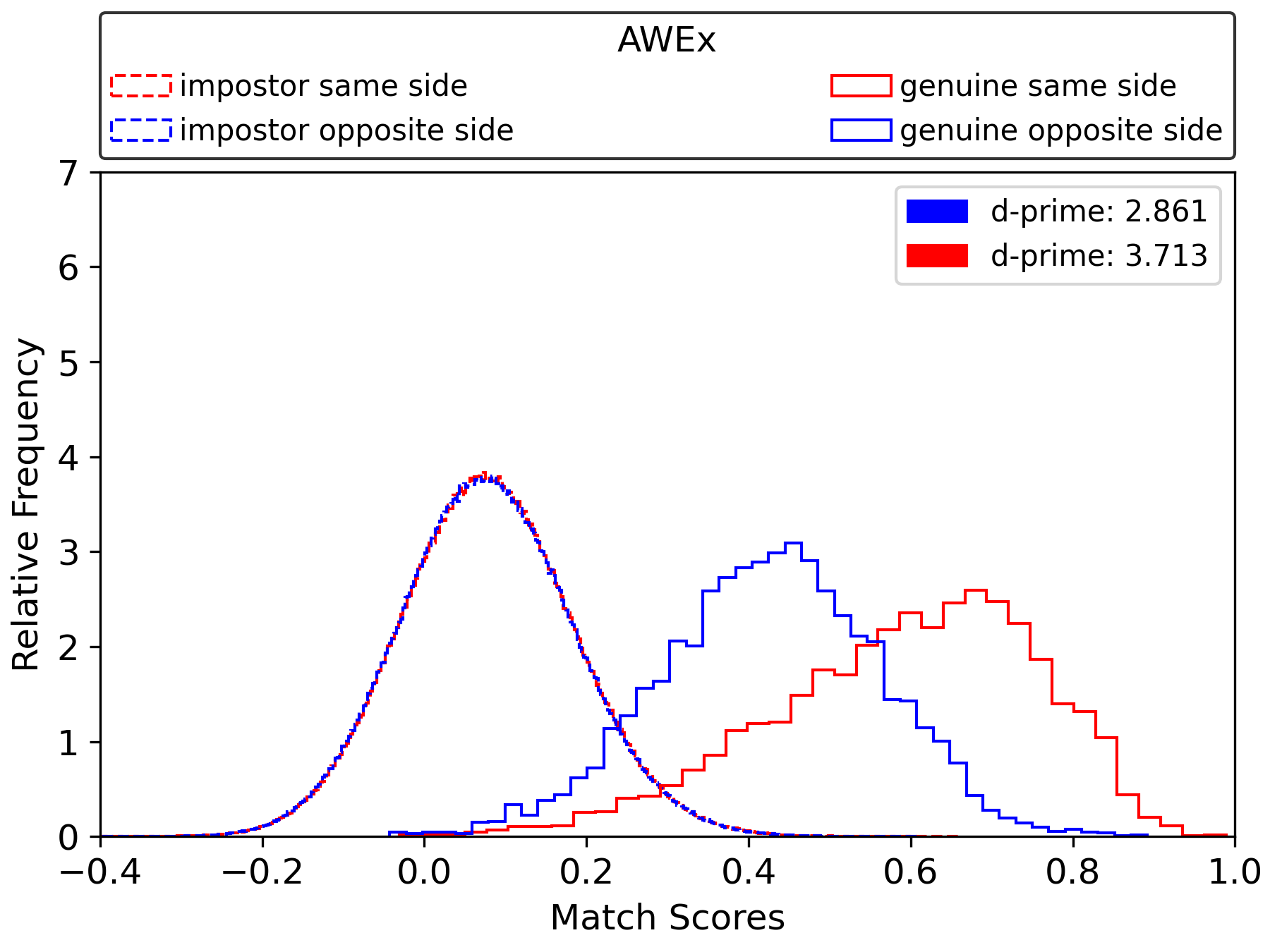}
        \label{fig:awe-ex}
        \vspace{-10pt}
        \caption{AWEx}
    \end{subfigure}

    \vspace{15pt} %

    \begin{subfigure}[b]{0.49\textwidth}
        \includegraphics[width=\textwidth]{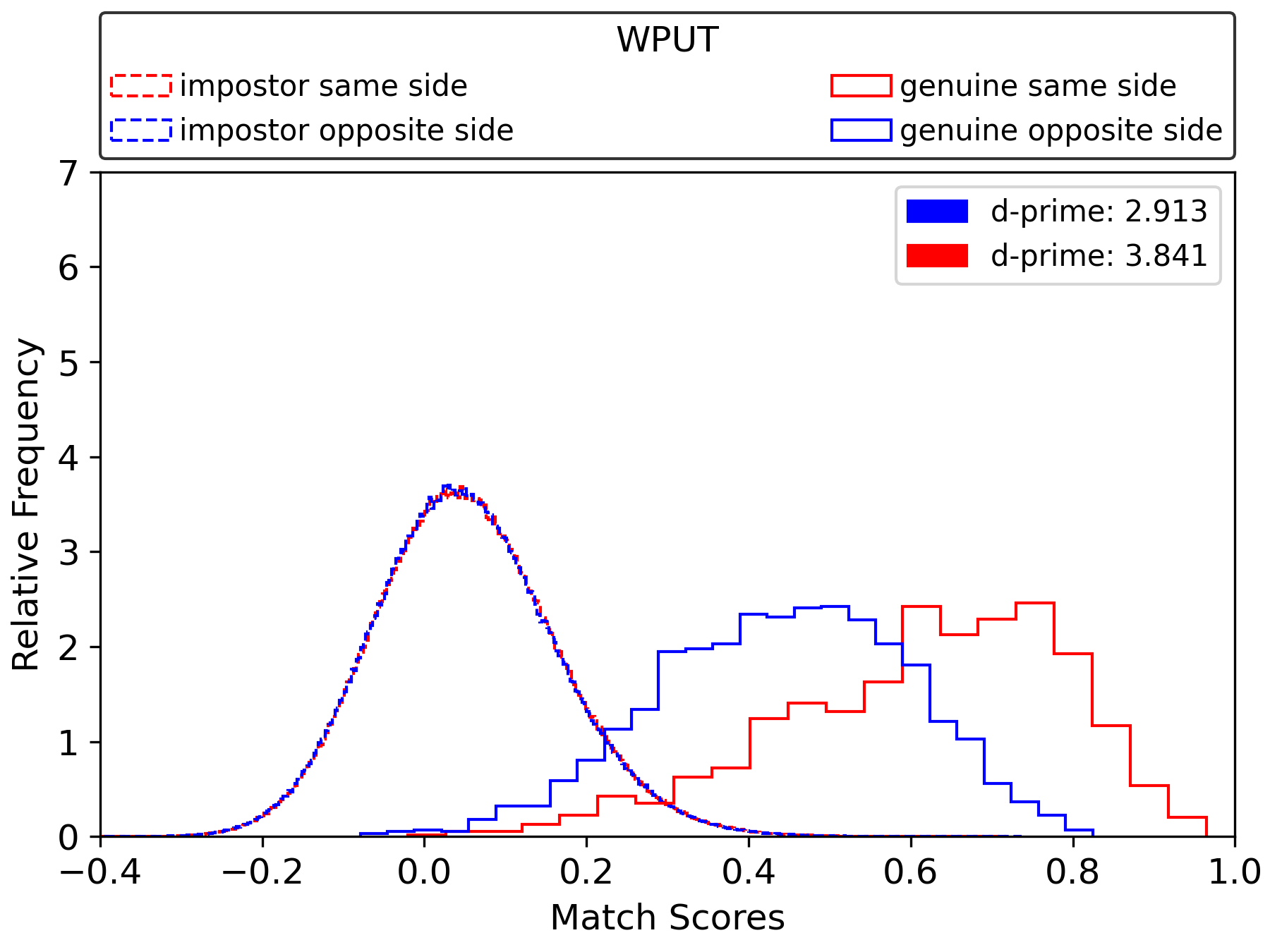}
        \label{fig:wput}
        \vspace{-10pt}
        \caption{WPUT}
    \end{subfigure}
    \hfill
    \begin{subfigure}[b]{0.49\textwidth}
        \includegraphics[width=\textwidth]{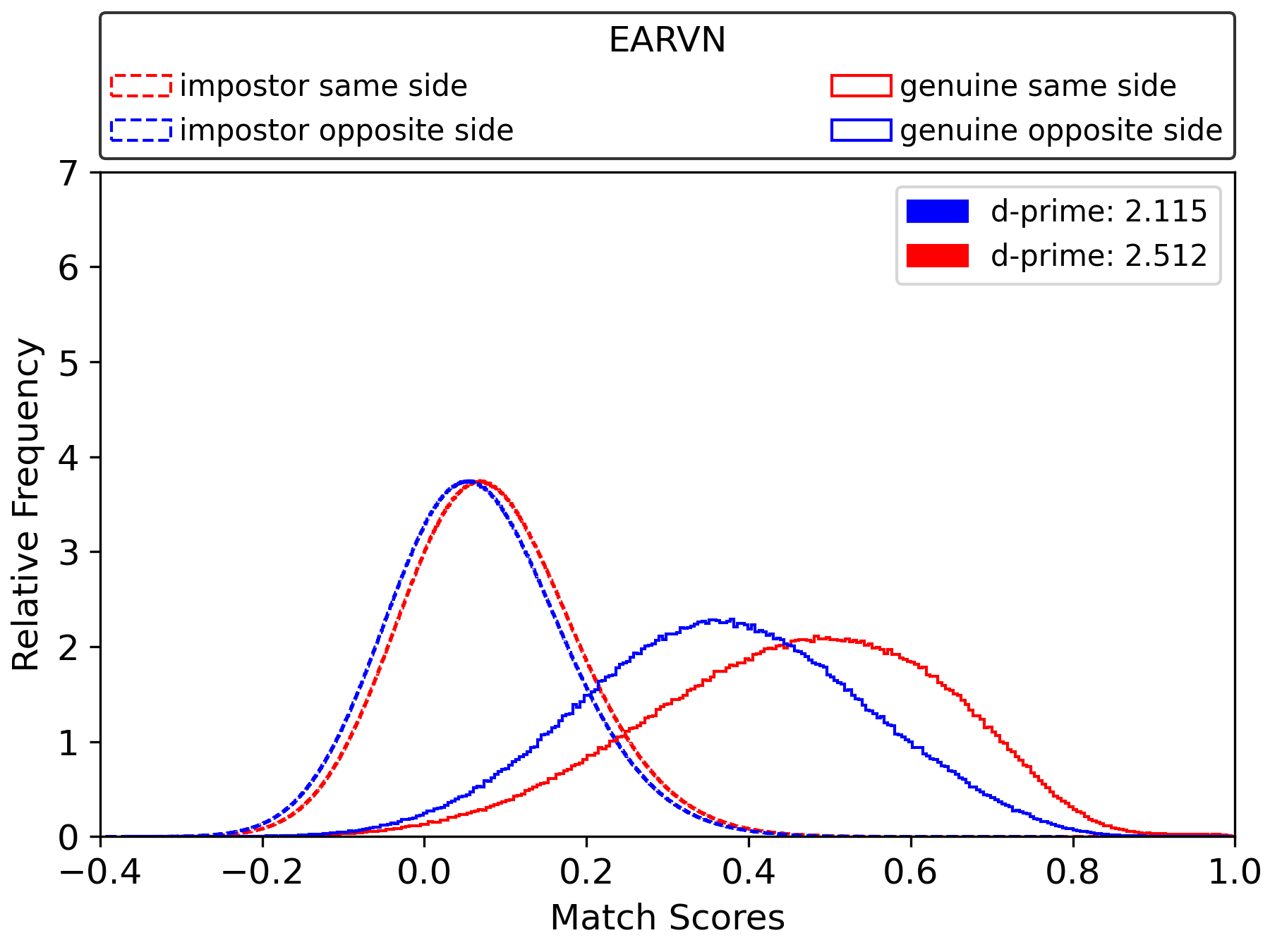}
        \label{fig:earvn}
        \vspace{-10pt}
        \caption{EARVN}
    \end{subfigure}
    
    \caption{Comparison of two matching strategies: same-side ear matching (left-to-left and right-to-right, shown in red) and opposite-side matching (left-to-right, shown in blue). Genuine and impostor score distributions are given for 4 datasets, with corresponding d-prime values shown in the top-right corner of each cell to indicate the separation between the distributions. Higher d-prime values show better separation between genuine and impostor distributions. While impostor score distributions remain consistent across both settings, genuine scores for opposite-side matching show a noticeable shift toward lower similarity values, suggesting reduced verification performance.}
    \label{fig:plots-ear-side}
\end{figure*}

We next evaluate the impact of the proposed alignment strategy on ear recognition performance. As illustrated in Figure~\ref{fig:alignment}, the method begins by detecting the ear region within a cropped ear image. The top-$k$ longest lines within the segmented area are then identified to estimate the top and bottom points of the ear. To reduce computational complexity, images are initially resized such that their smaller dimension is set to 80 pixels while preserving the aspect ratio. The top-50 longest lines are used in our experiments to enhance the robustness of the alignment estimation. Results for four alignment configurations are presented in Table~\ref{tab:align}. While the best average performance is achieved without applying alignment during either training or testing, the highest performance on the challenging EarVN dataset is observed when the proposed alignment strategy is employed. These findings suggest that, although alignment may not be critical for tightly cropped ear images, it can significantly improve recognition accuracy in more challenging scenarios, such as those found in the EarVN dataset, which includes greater variability in pose and resolution compared to the other benchmarks.

Lastly, we investigate the effect of left–right ear similarity on ear recognition performance. The results are given in Table~\ref{tab:side}, where we assess the impact of ear symmetry in both training and testing strategies. For training, two ear recognition models are developed, as illustrated in Figure~\ref{fig:training}: (i) a model with a single class per subject, and (ii) a model with two separate classes per subject to distinguish between left and right ears. During testing, two protocols are employed, as shown in Figure~\ref{fig:test}, to measure verification performance. Results indicate that better performance is achieved when verification protocol is restricted to ears from the same side (\ie , left-to-left or right-to-right). This suggests that CNN-based representations encode distinct features for left and right ears, regardless of whether the model is explicitly trained to separate them. This asymmetry is further illustrated in Figure~\ref{fig:plots-ear-side}, where the distribution of impostor scores remains similar across same-side and opposite-side comparisons, while genuine similarity scores are noticeably lower when matching opposite-side ears. 

Our analysis on the effect of incorporating side information during training suggest that, for a test scenario where a system need to perform verification with opposite-side ears, it is beneficial to train a model that considers left and right ears as a single class. On the other hand, if matching ears are of the same side, verification rates can be improved with treating left and right ears as separate classes during training. Note, the degree of symmetry can vary between people and evaluation on more subjects is required for more rigorous conclusions.

\textbf{Comparison with Previous Work.}
Table~\ref{tab:sota} presents a performance comparison between our method and prior work on the IITD-II dataset~\cite{kumar2012automated}. Note, unlike previous approaches which use a train-test split with overlapping subjects across both sets, our method is evaluated in a cross-dataset setting, without using any IITD-II images during training.

\section{Conclusion}
In this work, we analyze the impact of ear symmetry on recognition performance. Our experiments demonstrate that CNN-based representation learning can exploit bilateral ear asymmetry during training. Furthermore, we explore the effects of various configurations, such as ear alignment, input resolution and finetuning, on recognition accuracy. Cross-dataset evaluations reveal that despite the relatively limited size of the training set compared to face recognition datasets, ear images collected from existing face datasets can be effectively utilized to develop recognition models with strong generalization capabilities. High performance is achieved on three benchmark datasets, AWE, AWEx, and WPUT highlighting the potential of ear biometrics as an alternative to face recognition.

However, results on the EarVN dataset indicate that significant challenges remain in more difficult scenarios such as a large variation in pose and image quality. We encourage future research to incorporate EarVN into evaluation protocols, as it provides a valuable benchmark for ear recognition systems. Although EarVN has been used for performance evaluation in several works \cite{ramos2022vggfaceear, alshazly2020deep, alshazly2021towards}, train-test splits within the dataset is performed, limiting the ability to assess out-of-distribution performance of ear recognition approaches.

{
    \small
    \bibliographystyle{ieeenat_fullname}
    \bibliography{main}
}

\end{document}